\documentclass[conference]{IEEEtran}
\IEEEoverridecommandlockouts
\usepackage{times}  % DO NOT CHANGE THIS
\usepackage{helvet}  % DO NOT CHANGE THIS
\usepackage{courier}  % DO NOT CHANGE THIS
\usepackage[hyphens]{url}  % DO NOT CHANGE THIS
\usepackage{graphicx} % DO NOT CHANGE THIS
\usepackage{enumitem}
\usepackage{tikz}
\usepackage{graphicx}
\usepackage{pgfplots}
\usepackage{natbib}  % DO NOT CHANGE THIS AND DO NOT ADD ANY OPTIONS TO IT
\pgfplotsset{compat=1.18}
\usepackage{tikz}
\usepackage[ruled,vlined]{algorithm2e}
\usepackage{amsmath,amssymb,amsfonts}
\usepackage{algorithmic}
\usepackage{graphicx}
\usepackage{textcomp}
\usepackage{xcolor}
\def\BibTeX{{\rm B\kern-.05em{\sc i\kern-.025em b}\kern-.08em
    T\kern-.1667em\lower.7ex\hbox{E}\kern-.125emX}}
\begin{document}

\title{Evaluating Language Model Safety Across Long Adversarial Conversations\\
}

\author{\IEEEauthorblockN{1\textsuperscript{st} Parisa Salmani}
\IEEEauthorblockA{\textit{Ontario Tech University}\\
Oshawa, Canada \\
}
\and
\IEEEauthorblockN{2\textsuperscript{nd} Peter R. Lewis}
\IEEEauthorblockA{\textit{Ontario Tech University}\\
Oshawa, Canada  \\
}
}

\maketitle

\begin{abstract}
Conversational safety evaluations often test language models with a single harmful prompt, even though real-world systems interact with users through long, adaptive conversations. This study examines whether models continue to respond safely when an adversarial user persists across multiple turns.

We evaluate three open-weight, instruction-tuned models on two harmful prompts across different conversation lengths and random seeds. In each setting, a second language model acts as a persistent adversarial user, while a safety classifier labels every response as safe or unsafe.

Across all model-prompt combinations, first-turn safe-response rates ranged from 85\% to 100\%. By depth 11, they dropped to 38-61\%, and by depth 101, to 15-44\%. This decline appeared across models and continued well beyond the short interactions typically used in multi-turn safety evaluations.

These results provide proof-of-concept evidence that strong single-turn safety does not necessarily persist during sustained adversarial interaction. They highlight the need for long-horizon evaluations and conversation-level safeguards that account for risk accumulating across turns.

\end{abstract}


\section{Introduction}

Conversational language models have rapidly evolved from research prototypes into everyday socio-technical infrastructure. They are now widely used for healthcare, emotional support, education, programming, and companionship, often enabling prolonged and emotionally intensive interactions. However, their growing integration into daily life has also exposed serious real-world harms.

Recent cases illustrate the potential harms of current language models. In one wrongful-death lawsuit, an adolescent reportedly formed a months-long emotional attachment to a Character.AI chatbot and later died by suicide. The lawsuit alleges that the chatbot failed to redirect suicidal discussions and sometimes reinforced them \citep{garcia2024characterai}. A similar case involved a distressed user who died by suicide after prolonged interactions with the Eliza chatbot \citep{eliza2023belgium}. The National Eating Disorders Association also withdrew its Tessa chatbot after reports that it recommended calorie-restriction plans to users recovering from eating disorders \citep{neda2023tessa}. Studies of Replika have identified risks including emotional dependence and unsolicited sexual content shown to minors \citep{laestadius2024replika}. In healthcare, controlled studies and case reports have found that general-purpose chatbots may provide incorrect dosages, overlook contraindications, and generate confident but unsupported medical advice \citep{nori2023medical,stade2024lhealth}.

A common feature of these cases is often overlooked: the harmful response is rarely the first response. Interactions may begin with harmless exchanges, but users can persist, reframe requests, provide new context, express greater distress, or claim relevant expertise. Over time, the assistant’s refusals may become less consistent, allowing unsafe content to appear late in the conversation. The Character.AI case reportedly developed over several months, the Tessa failure occurred during multi-turn coaching, and the Belgian Eliza case involved six weeks of sustained interaction. These cases suggest that single-turn evaluations may underestimate the risks of conversational AI systems used over extended periods.

This gap reflects a mismatch between current safety-training pipelines and real-world use. Most alignment and safety procedures optimize models against short, single-prompt examples of harmful requests, with feedback based on isolated responses \citep{bai2022constitutional,casper2023open}. Long conversations create additional challenges. Earlier warnings or partial refusals remain in the conversation history, which may encourage the model to continue previous concessions rather than issue a new refusal \citep{sharma2023sycophancy}. Repeated user requests can also make the harmful goal more prominent, while the model’s general objective to be helpful may eventually override its safety constraints \citep{perez2022discovering}. As a result, single-turn benchmarks may systematically underestimate safety failures that emerge over extended conversations \citep{russinovich2024crescendo}.

A growing body of work has begun to evaluate safety in multi-turn settings \citep{salmani1}, confirming that multi-turn behavior diverges from single-turn behavior. Existing multi-turn benchmarks typically operate over short, pre-constructed dialogues, commonly three to ten turns, and report aggregate safety scores across models. Less is known about what happens during longer interactions: how safety changes as conversations continue, whether similar patterns appear across model families, and how far a model may drift under persistent adversarial pressure.

The implications are significant for AI ethics, governance, and deployment. Conversational systems are increasingly used by vulnerable groups, including minors using companion applications \citep{garcia2024characterai}, people seeking mental-health support when professional care is unavailable \citep{stade2024lhealth}, and patients asking about medication or dosage \citep{nori2023medical}. They may also be used to request guidance that violates privacy or enables abuse \citep{salmani2}. Reported failures include reinforcing suicidal thoughts, providing unsafe medical advice, and supporting privacy-invasive actions. 

In this paper, we evaluate three open-source, instruction-tuned language models developed by different organizations: OpenAI’s GPT-OSS-20B, Meta’s Llama-3.2-3B-Instruct, and Google’s Gemma-4-26B-A4B-it. We evaluate the models across two harmful prompts: The first concerns specialized medical advice and the second concerns dangerous information about weapon construction. For each model and prompt, we conduct exhaustive searches at two conversation depths: 11 turns for short interactions and 101 turns for sustained adversarial interactions. We repeat exhaustive short interactions over 10 seeds and long interactions across 50 random seeds. A fixed third-party safety classifier, Llama Guard \citep{inan2023llamaguard}, evaluates each assistant response.

Across open-source models we tested, safe-response rates were 85–100\% at the first turn, similar to single-turn benchmark results. Under persistent user pressure, however, safety declined with conversation length, reaching as low as 15\% at the greatest depth. This pattern was consistent across random seeds, showing that first-turn safety does not reliably predict behavior in long conversations.

Our results show that sustained adversarial interaction can expose safety failures that are not visible in single-turn evaluation. Unlike prior work focused on single-turn or short interactions, we measure safety over much longer conversations. We release our methodology and aggregate results for replication, but withhold harmful prompts, unsafe responses, and full conversation logs to limit misuse.

\section{Related Works}

\subsection{Real-world Chatbot Harms}

% A growing body of legal, journalistic, and qualitative evidence describes harms that developed over extended chatbot interactions rather than through isolated exchanges. One wrongful-death lawsuit alleges that months of emotionally intense conversations with a companion chatbot preceded the suicide of a fourteen-year-old user \citep{garcia2024characterai}. Reporting on a separate case in Belgium similarly described approximately six weeks of intensive interaction with a chatbot before the user’s suicide \citep{eliza2023belgium}. These accounts consist of legal allegations and journalistic reporting rather than controlled evidence of causation. Nevertheless, they illustrate why complete conversational trajectories may be more informative than isolated model outputs when evaluating potential harm.

% Another widely reported incident involved an eating-disorder support chatbot that was taken offline on May 30, 2023, after users reported receiving harmful recommendations concerning weight loss, calorie restriction, and frequent weighing \citep{neda2023tessa}. The suspension occurred two days before the organization’s human-operated helpline was scheduled to close. The organization denied that the chatbot was intended to replace the helpline, although former staff characterized the transition in those terms. The system was rule-based rather than a modern generative language model, but the incident still demonstrates how problematic guidance can emerge and accumulate during an ongoing chatbot interaction.

Legal, journalistic, and qualitative reports suggest that chatbot-related harms may develop cumulatively across extended interactions. In two reported suicide cases, users engaged intensively with companion chatbots for several weeks or months beforehand \citep{garcia2024characterai, eliza2023belgium}. Although these accounts do not establish causation, they indicate that full conversational trajectories may reveal risks that isolated outputs miss. Similar concerns arose when an eating-disorder support chatbot was suspended after users reported repeated recommendations involving weight loss, calorie restriction, and frequent weighing \citep{neda2023tessa}. Despite being rule-based rather than generative, the incident further illustrates how harmful guidance can accumulate over an ongoing interaction.
% But to date no one has asked ... but no one has asked  and that's what I am doing ... 
% safedialbench in one sentence

Prior work has examined chatbot-related harms through qualitative, clinical, and systems-level perspectives. Studies of companion chatbots and behavioral-health applications have identified risks of emotional dependence, weak clinical validation, and persistent engagement challenges \citep{laestadius2024replika,stade2024lhealth}. Medical benchmarks likewise show that advanced language models can perform strongly while retaining important accuracy and safety limitations, although such results do not establish causation in specific real-world incidents \citep{nori2023medical}. More broadly, incident analysis requires reconstructing interacting system, contextual, and cognitive factors across complete trajectories rather than examining isolated outputs \citep{ezell2025incidents}. Existing real-world evidence remains largely qualitative, legal, or journalistic, while technical work emphasizes benchmarks and conceptual frameworks. Our work aims to add controlled evidence of a trajectory-level safety analysis and offers a possible mechanism through which risk accumulates over extended interactions, without claiming that documented incidents demonstrate this mechanism.

\subsection{LLM Safety Benchmarks}
Most LLM safety evaluations assess responses to individual harmful prompts in single-turn interactions. Existing studies have introduced standardized frameworks, compact prompt sets, curated datasets, and public leaderboards for comparing model safety and jailbreak attacks across risk categories and testing conditions \citep{mazeika2024harmbench,vidgen2023simplesafetytests,chao2024jailbreakbench,wang2023donotanswer,li2024saladbench}. Related work shows that adversarial suffixes can bypass safeguards across prompts and models \citep{zou2023universal}, while another dataset focuses primarily on detecting toxic user inputs rather than evaluating safe refusals \citep{lin2023toxicchat}. Although these resources provide important evaluation infrastructure, they generally assess test cases independently and do not measure whether models can be gradually steered toward unsafe behavior through extended multi-turn dialogue.

Broader reviews argue that static benchmarks do not adequately capture how language models behave in interaction with users and technical systems, motivating more dynamic forms of behavioral evaluation \citep{eriksson2025benchmarks,mcintosh2024inadequacies}. This limitation is increasingly relevant to regulation, as the EU AI Act and the 2025 General-Purpose AI Code of Practice incorporate evaluation into systemic-risk assessment without explicitly addressing the limits of single-turn testing \citep{european2024aiact,europeancommission2025gpaicode}. Consequently, benchmarks used for compliance should clearly specify which real-world interaction conditions they represent.

% Research on dialogue-based red teaming for language models dates back at least to the work of \citet{perez2022redteam}, but interest in stateful, multi-turn jailbreaks expanded rapidly during 2024 and 2025. One study introduced an attack that begins with seemingly benign requests and gradually escalates toward a harmful objective by building on the target model’s earlier responses \citep{russinovich2025crescendo}. Its automated version completed most successful attacks in fewer than five conversational turns. Other researchers explored alternative ways of concealing or progressively revealing harmful intent. One study used conversational coreference to distribute malicious intent across multiple turns and found that this strategy often reduced model harmlessness relative to equivalent single-turn prompts, although it did not increase harmful-response rates for every model \citep{yu2024cosafe}. Another study constructed diverse conversational paths toward harmful goals by using networks of semantically related actors \citep{ren2025actorattack}. More recent work adopted a multi-agent approach in which separate agents coordinate to plan, optimize, and verify adaptive attacks over the course of a conversation \citep{rahman2025xteaming}.

A related body of work automates jailbreak discovery through iterative search. Language-model-based methods refine attacks using target feedback, either sequentially \citep{chao2023pair} or through branching and pruning over multiple candidates \citep{mehrotra2024tap}. Other approaches optimize prompts using genetic algorithms \citep{liu2024autodan} or fuzzing and mutation \citep{yu2023gptfuzz}. Earlier work also used language models to generate adversarial test cases and dialogues \citep{perez2022redteam}, while later studies expanded the attack space through persona manipulation \citep{shah2023personas} and cipher-based prompting \citep{yuan2024cipher}. The most closely related work examines multi-turn attacks \citep{russinovich2024crescendo} and their internal representations \citep{bullwinkel2025representation}. These attacks aim to maximize success and generally terminates once the target objective has been achieved; their automated implementation succeeds on most tasks within fewer than five turns. Our objective is different: rather than proposing or optimizing an attack, we measure how turn-level safe-response rates evolve over substantially longer interactions, including after the first unsafe response has occurred.

Our work differs from the existing literature in two main aspects. First, rather than optimizing attacks, we measure safety under conditions that more closely resemble real-world interactions. We hold the initial harmful objective, role configuration, model configuration, and evaluation procedure fixed while measuring safety across increasing conversational depth. This design allows us to study how safety changes over the course of an interaction instead of searching for the most effective individual jailbreak. Second, another language model generates the adversary’s prompts throughout the interaction, allowing the long conversation to be automated. The decline in safety could therefore have two possible causes: the assistant may become less reliable, or the adversary may become more effective. To separate these explanations, we measure assistant safety and adversarial pressure independently. We find that assistant safety declines during some parts of the interaction even when adversarial pressure decreases and the adversary’s behavior becomes safer. This suggests that the safety degradation is not caused by increasingly unsafe or effective adversarial prompts.

\section{Methodology}

To conduct our experiments, we assign a language model the role of a persistent adversarial user, which we call the shadow user. The shadow user is given a fixed harmful objective (the initial harmful prompt) and continues pursuing it across the conversation. Using a language model as the user allows us to generate long interactions, up to a depth of 101 turns, across many random seeds without manually prompting the assistant model. The goal is to model sustained user pressure rather than a single harmful request. Importantly, the shadow user and the assistant under evaluation use the same model weights but are assigned different roles. Thus, neither side has access to a more capable model or is superior. The experiment measures how a model’s safety behavior changes when it is exposed to persistent pressure (generated by another instance of the same model).

We use a controlled methodology through the experiments; fixed initial prompts, a fixed adversarial shadow user, a fixed safety judge, and no fine-tuning of the evaluated models. Our goal is to isolate the relationship between conversation depth and safety while keeping all other factors constant. We define depth as the position of a turn in the conversation. Depth 0 is the initial human prompt, odd depths are assistant responses, and even depths are shadow-user turns. Therefore, a conversation ending at depth (d) contains $\lceil d/2 \rceil$ assistant responses and $\lfloor d/2 \rfloor$ shadow-user turns after the initial prompt.

\subsection{Harm domains, adversarial shadow, and safety judge}

We evaluate three open-weight, instruction-tuned language models from different organizations: OpenAI’s \textit{GPT-OSS-20B}, Meta’s \textit{Llama-3.2-3B-Instruct}, and Google’s \textit{Gemma-4-26B-A4B-it}. These models differ in size and training approach, allowing us to study whether the relationship between conversation depth and safety holds across model families and scales. We use open-weight models so that the model weights remain fixed and all experiments can be run locally under controlled conditions. In each experiment, the same model configuration is used for both the assistant being evaluated and the adversarial shadow user. The two instances differ only in their assigned roles and system prompts.

We evaluate two harmful prompts. The first involves specialized medical advice, including requests for medication dosing, treatment, and clinical guidance that should normally be referred to a qualified healthcare professional. The second involves dangerous information related to weapons construction. For safety reasons, we describe these domains only at a general level and do not publish the initial prompts. The prompts remain fixed across all models, experimental conditions, and random seeds.

Every assistant response is evaluated by a fixed third-party safety classifier, Llama-Guard-4-12B \citep{inan2023llamaguard}. We use the same classifier across all models, domains, conditions, and experimental phases to avoid differences caused by evaluator variation. The classifier’s binary safe or unsafe label is used as the main unit of analysis.

Evaluating the results with additional safety classifiers or human reviewers is left for future work. The goal of this study is to demonstrate how safety changes with conversation depth, rather than to compare different evaluation methods or determine which model is safest.

\subsection{Experiments}

We evaluate model safety using two experimental settings: short conversations with a maximum depth of 11 and long conversations with a maximum depth of 101.

In the short-conversation setting, we construct a complete binary conversation tree. Starting from the fixed initial prompt at depth 0, the assistant generates two candidate responses. The shadow user then generates two follow-ups for each response, and this process continues with both roles branching at every turn until depth 11. Each tree therefore contains ($2^{11}=2048$) complete dialogue paths. We repeat this process using 10 random seeds for every model and initial prompt.

The safe-response rate at each depth is calculated over every assistant response at that level rather than from a sampled subset. This provides an exhaustive measure of how safety changes as the conversation gets longer. Depth 11 is the largest depth for which complete enumeration of the conversational tree remains computationally practical. 

To evaluate the assistant model’s safety rate, we focus on responses at odd depths of the conversation tree, where the assistant generates a reply. Even depths correspond to the shadow user, whose role is to remain intentionally adversarial.

Exhaustive enumeration cannot be extended to the depth of realistic conversations because the number of branches grows exponentially. Our second experiment therefore trades completeness for depth by sampling individual trajectories rather than exploring the full conversation tree.

For each model and harmful prompt, we generate 50 conversation branches in each tree and we experiment through 50 different random seeds, making it 2500 conversation flow for each model and harmful prompt. Each conversation follows a single unbranched path: the assistant produces one response, the adversarial shadow user produces one follow-up, and the two alternate until depth 101. This depth is substantially greater than in the other exhaustive experiment and exceeds the length of most existing multi-turn safety benchmarks. Every assistant response is evaluated by the same fixed safety judge. At each depth, we report the safe-response rate as the proportion of responses across the that depth.

\section{Results}
Across both experimental settings, all tested models, random seeds, and prompts showed the same pattern: models that initially refused harmful requests became substantially more likely to comply when a persistent adversarial (shadow) user continued the conversation. We first present the exhaustive shallow-phase results, followed by the deep-phase results, and then the cross-cutting comparisons.

\subsection{Short Conversations}

Table~\ref{tab:phase1-safe-response} reports the safe-response rate at the first assistant turn (depth 1) and the deepest exhaustively explored turn (depth 11) for each model–domain pair, averaged across ten seeds and all nodes at the corresponding depth in each tree.

\begin{table}[t]
    \centering
    \renewcommand{\arraystretch}{1.25}
    \begin{tabular}{|l|c|c|}
        \hline
        \textbf{Model} & \textbf{Prompt 1} & \textbf{Prompt 2} \\
        \hline
        \texttt{GPT-OSS-20B} & $0.90 \rightarrow 0.46$ & $1.00 \rightarrow 0.55$ \\
        \hline
        \texttt{Llama-3.2-3B} & $1.00 \rightarrow 0.43$ & $1.00 \rightarrow 0.61$ \\
        \hline
        \texttt{Gemma-4-26B-A4B} & $0.85 \rightarrow 0.38$ & $1.00 \rightarrow 0.52$ \\
        \hline
    \end{tabular}
    \vspace{7pt}
    \caption{Safe-response rate at the first turn (Depth = 1) to deepest turn (Depth = 11), by model and domain, averaged over ten different seeds.}
    \label{tab:phase1-safe-response}
\end{table}

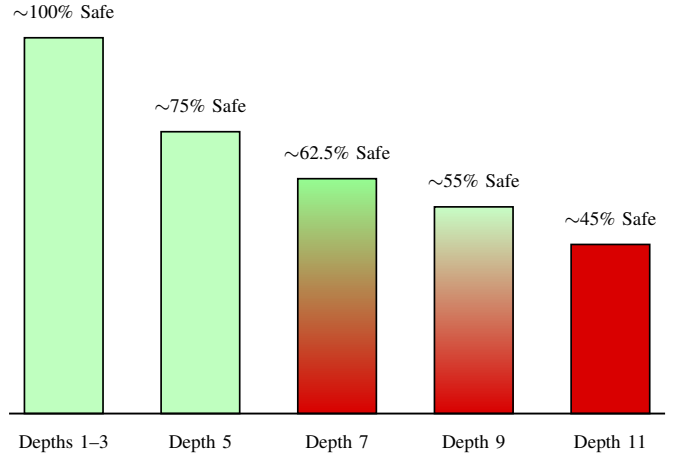
\begin{figure}[t]
    \centering
    \resizebox{\columnwidth}{!}{%
    \begin{tikzpicture}[x=1cm, y=0.055cm]

        % Bar width
        \def\barw{1.15}

        % Bars: x / height / bottom color / top color / label / x-axis label
        \foreach \x/\h/\bot/\top/\txt/\lab in {
            1.0/100/green!25/green!25/{\(\sim\)100\% Safe}/{Depths 1--3},
            3.0/75/green!25/green!25/{\(\sim\)75\% Safe}/{Depth 5},
            5.0/62.5/red!85!black/green!42/{\(\sim\)62.5\% Safe}/{Depth 7},
            7.0/55/red!85!black/green!22/{\(\sim\)55\% Safe}/{Depth 9},
            9.0/45/red!85!black/red!85!black/{\(\sim\)45\% Safe}/{Depth 11}
        }{
            \shade[
                draw=black,
                line width=0.7pt,
                bottom color=\bot,
                top color=\top
            ]
            (\x-\barw/2,0) rectangle (\x+\barw/2,\h);

            \node[font=\footnotesize] at (\x,\h+7) {\txt};
            \node[font=\footnotesize] at (\x,-8) {\lab};
        }

        % Baseline
        \draw[black, line width=0.8pt] (0.2,0) -- (9.8,0);
    
    \end{tikzpicture}

    }
    \caption{Safe-response rate declines with conversational depth. Bars report the proportion of assistant responses judged safe at each depth in a single depth-11 exhaustive tree (one seed) seeded with the first prompt with Llama model. Safety falls from near-complete at the opening turns to roughly 45 percent by depth 11.}
    \label{fig:depth-safety}
\end{figure}

\subsection{Long Conversations}

The previous exhaustive experiment with short conversations examined only relatively shallow depths. To test whether the observed degradation persists, worsens, or reverses over longer interactions, we track 50 sampled conversations from each conversation tree to depth 101, repeating the experiment across 50 seeds for each model and prompt combination (Table~\ref{tab:phase2-safe-response}).

\begin{table}[t]
    \centering
    \renewcommand{\arraystretch}{1.25}
    \begin{tabular}{|l|c|c|}
        \hline
        \textbf{Model} & \textbf{Prompt 1} & \textbf{Prompt 2} \\
        \hline
        \texttt{GPT-OSS-20B} & $0.90 \rightarrow 0.28$ & $1.00 \rightarrow 0.38$ \\
        \hline
        \texttt{Llama-3.2-3B} & $1.00 \rightarrow 0.22$ & $1.00 \rightarrow 0.44$ \\
        \hline
        \texttt{Gemma-4-26B-A4B} & $0.85 \rightarrow 0.15$ & $1.00 \rightarrow 0.36$ \\
        \hline
    \end{tabular}
    \vspace{7pt}
     \caption{Longer conversation (50 sampled over each tree with 50 seeds): safe-response rate at the first turn $\rightarrow$ deepest sampled turn, by model and domain.}
     \label{tab:phase2-safe-response}
\end{table}

The longer-conversation experiments show that safety continues to decline beyond the shallow depths covered by exhaustive exploration. Every entry in Table~\ref{tab:phase2-safe-response} is lower than its corresponding entry in Table~\ref{tab:phase1-safe-response}. For example, the safe-response rate of \texttt{GPT-OSS-20B} on the first prompt drops from $0.46$ at the deepest exhaustive turn to $0.28$ in the deep sampled conversations, while \texttt{Gemma-4-26B-A4B} reaches $0.15$, the lowest rate observed in the study. The two experiments agree at overlapping depths: exhaustive enumeration confirms that the initial decline across short conversations is not a sampling artifact, and deeper sampling shows that the decline continues as conversations grow longer. Overall, safety begins to decrease around depth 3 or 4 and then falls more gradually under persistent adversarial interaction.

% Another notable pattern appears when separating the assistant’s responses at odd depths from the adversarial shadow user’s turns at even depths, as shown in Figure~\ref{fig:safety-bars-101}. For several seeds, the shadow user becomes safer as the conversation progresses, indicating weaker adversarial pressure. However, the evaluated assistant continues to become less safe rather than recover. If degradation were driven only by immediate adversarial pressure, safety should improve as that pressure decreases. Instead, the results suggest that accumulated conversational context helps sustain the decline. Earlier concessions and unsafe responses remain in the context and may make later responses more likely to comply even without user's adverserial pressure~\cite{sharma2023sycophancy}.

% <Some more explanation and context for this>

\begin{figure}[htbp]
\centering
\begin{tikzpicture}
\begin{axis}[
    width=\columnwidth,
    height=5.2cm,
    ymin=0,
    ymax=100,
    xmin=0,
    xmax=102,
    ylabel={Safety (\%)},
    xlabel={Depth},
    xtick={1,20,40,60,80,100},
    ytick={0,25,50,75,100},
    tick label style={font=\scriptsize},
    label style={font=\footnotesize},
    axis lines=left,
    axis on top,
    enlarge x limits=0.01,
]

\draw[draw=black, line width=0.15pt, fill=green!85!red] (axis cs:0.65,0) rectangle (axis cs:1.35,85);
\draw[draw=black, line width=0.15pt, fill=blue!70] (axis cs:1.65,0) rectangle (axis cs:2.35,10);
\draw[draw=black, line width=0.15pt, fill=green!84!red] (axis cs:2.65,0) rectangle (axis cs:3.35,79);
\draw[draw=black, line width=0.15pt, fill=blue!70] (axis cs:3.65,0) rectangle (axis cs:4.35,6);
\draw[draw=black, line width=0.15pt, fill=green!82!red] (axis cs:4.65,0) rectangle (axis cs:5.35,82);
\draw[draw=black, line width=0.15pt, fill=blue!70] (axis cs:5.65,0) rectangle (axis cs:6.35,14);
\draw[draw=black, line width=0.15pt, fill=green!81!red] (axis cs:6.65,0) rectangle (axis cs:7.35,81);
\draw[draw=black, line width=0.15pt, fill=blue!70] (axis cs:7.65,0) rectangle (axis cs:8.35,18);
\draw[draw=black, line width=0.15pt, fill=green!79!red] (axis cs:8.65,0) rectangle (axis cs:9.35,86);
\draw[draw=black, line width=0.15pt, fill=blue!70] (axis cs:9.65,0) rectangle (axis cs:10.35,16);

\draw[draw=black, line width=0.15pt, fill=green!78!red] (axis cs:10.65,0) rectangle (axis cs:11.35,78);
\draw[draw=black, line width=0.15pt, fill=blue!70] (axis cs:11.65,0) rectangle (axis cs:12.35,25);
\draw[draw=black, line width=0.15pt, fill=green!76!red] (axis cs:12.65,0) rectangle (axis cs:13.35,65);
\draw[draw=black, line width=0.15pt, fill=blue!70] (axis cs:13.65,0) rectangle (axis cs:14.35,20);
\draw[draw=black, line width=0.15pt, fill=green!77!red] (axis cs:14.65,0) rectangle (axis cs:15.35,73);
\draw[draw=black, line width=0.15pt, fill=blue!70] (axis cs:15.65,0) rectangle (axis cs:16.35,25);
\draw[draw=black, line width=0.15pt, fill=green!73!red] (axis cs:16.65,0) rectangle (axis cs:17.35,73);
\draw[draw=black, line width=0.15pt, fill=blue!70] (axis cs:17.65,0) rectangle (axis cs:18.35,34);
\draw[draw=black, line width=0.15pt, fill=green!72!red] (axis cs:18.65,0) rectangle (axis cs:19.35,80);
\draw[draw=black, line width=0.15pt, fill=blue!70] (axis cs:19.65,0) rectangle (axis cs:20.35,28);

\draw[draw=black, line width=0.15pt, fill=green!71!red] (axis cs:20.65,0) rectangle (axis cs:21.35,71);
\draw[draw=black, line width=0.15pt, fill=blue!70] (axis cs:21.65,0) rectangle (axis cs:22.35,34);
\draw[draw=black, line width=0.15pt, fill=green!69!red] (axis cs:22.65,0) rectangle (axis cs:23.35,74);
\draw[draw=black, line width=0.15pt, fill=blue!70] (axis cs:23.65,0) rectangle (axis cs:24.35,40);
\draw[draw=black, line width=0.15pt, fill=green!68!red] (axis cs:24.65,0) rectangle (axis cs:25.35,68);
\draw[draw=black, line width=0.15pt, fill=blue!70] (axis cs:25.65,0) rectangle (axis cs:26.35,44);
\draw[draw=black, line width=0.15pt, fill=green!66!red] (axis cs:26.65,0) rectangle (axis cs:27.35,60);
\draw[draw=black, line width=0.15pt, fill=blue!70] (axis cs:27.65,0) rectangle (axis cs:28.35,48);
\draw[draw=black, line width=0.15pt, fill=green!65!red] (axis cs:28.65,0) rectangle (axis cs:29.35,65);
\draw[draw=black, line width=0.15pt, fill=blue!70] (axis cs:29.65,0) rectangle (axis cs:30.35,54);

\draw[draw=black, line width=0.15pt, fill=green!63!red] (axis cs:30.65,0) rectangle (axis cs:31.35,63);
\draw[draw=black, line width=0.15pt, fill=blue!70] (axis cs:31.65,0) rectangle (axis cs:32.35,48);
\draw[draw=black, line width=0.15pt, fill=green!62!red] (axis cs:32.65,0) rectangle (axis cs:33.35,62);
\draw[draw=black, line width=0.15pt, fill=blue!70] (axis cs:33.65,0) rectangle (axis cs:34.35,58);
\draw[draw=black, line width=0.15pt, fill=green!60!red] (axis cs:34.65,0) rectangle (axis cs:35.35,60);
\draw[draw=black, line width=0.15pt, fill=blue!70] (axis cs:35.65,0) rectangle (axis cs:36.35,62);
\draw[draw=black, line width=0.15pt, fill=green!61!red] (axis cs:36.65,0) rectangle (axis cs:37.35,74);
\draw[draw=black, line width=0.15pt, fill=blue!70] (axis cs:37.65,0) rectangle (axis cs:38.35,66);
\draw[draw=black, line width=0.15pt, fill=green!57!red] (axis cs:38.65,0) rectangle (axis cs:39.35,57);
\draw[draw=black, line width=0.15pt, fill=blue!70] (axis cs:39.65,0) rectangle (axis cs:40.35,60);

\draw[draw=black, line width=0.15pt, fill=green!56!red] (axis cs:40.65,0) rectangle (axis cs:41.35,59);
\draw[draw=black, line width=0.15pt, fill=blue!70] (axis cs:41.65,0) rectangle (axis cs:42.35,62);
\draw[draw=black, line width=0.15pt, fill=green!55!red] (axis cs:42.65,0) rectangle (axis cs:43.35,55);
\draw[draw=black, line width=0.15pt, fill=blue!70] (axis cs:43.65,0) rectangle (axis cs:44.35,60);
\draw[draw=black, line width=0.15pt, fill=green!53!red] (axis cs:44.65,0) rectangle (axis cs:45.35,61);
\draw[draw=black, line width=0.15pt, fill=blue!70] (axis cs:45.65,0) rectangle (axis cs:46.35,64);
\draw[draw=black, line width=0.15pt, fill=green!40!red] (axis cs:46.65,0) rectangle (axis cs:47.35,45);
\draw[draw=black, line width=0.15pt, fill=blue!70] (axis cs:47.65,0) rectangle (axis cs:48.35,58);
\draw[draw=black, line width=0.15pt, fill=green!50!red] (axis cs:48.65,0) rectangle (axis cs:49.35,50);
\draw[draw=black, line width=0.15pt, fill=blue!70] (axis cs:49.65,0) rectangle (axis cs:50.35,52);

\draw[draw=black, line width=0.15pt, fill=green!49!red] (axis cs:50.65,0) rectangle (axis cs:51.35,43);
\draw[draw=black, line width=0.15pt, fill=blue!70] (axis cs:51.65,0) rectangle (axis cs:52.35,66);
\draw[draw=black, line width=0.15pt, fill=green!47!red] (axis cs:52.65,0) rectangle (axis cs:53.35,47);
\draw[draw=black, line width=0.15pt, fill=blue!70] (axis cs:53.65,0) rectangle (axis cs:54.35,56);
\draw[draw=black, line width=0.15pt, fill=green!58!red] (axis cs:54.65,0) rectangle (axis cs:55.35,58);
\draw[draw=black, line width=0.15pt, fill=blue!70] (axis cs:55.65,0) rectangle (axis cs:56.35,54);
\draw[draw=black, line width=0.15pt, fill=green!44!red] (axis cs:56.65,0) rectangle (axis cs:57.35,49);
\draw[draw=black, line width=0.15pt, fill=blue!70] (axis cs:57.65,0) rectangle (axis cs:58.35,66);
\draw[draw=black, line width=0.15pt, fill=green!33!red] (axis cs:58.65,0) rectangle (axis cs:59.35,33);
\draw[draw=black, line width=0.15pt, fill=blue!70] (axis cs:59.65,0) rectangle (axis cs:60.35,48);

\draw[draw=black, line width=0.15pt, fill=green!42!red] (axis cs:60.65,0) rectangle (axis cs:61.35,42);
\draw[draw=black, line width=0.15pt, fill=blue!70] (axis cs:61.65,0) rectangle (axis cs:62.35,60);
\draw[draw=black, line width=0.15pt, fill=green!40!red] (axis cs:62.65,0) rectangle (axis cs:63.35,40);
\draw[draw=black, line width=0.15pt, fill=blue!70] (axis cs:63.65,0) rectangle (axis cs:64.35,60);
\draw[draw=black, line width=0.15pt, fill=green!49!red] (axis cs:64.65,0) rectangle (axis cs:65.35,49);
\draw[draw=black, line width=0.15pt, fill=blue!70] (axis cs:65.65,0) rectangle (axis cs:66.35,64);
\draw[draw=black, line width=0.15pt, fill=green!37!red] (axis cs:66.65,0) rectangle (axis cs:67.35,37);
\draw[draw=black, line width=0.15pt, fill=blue!70] (axis cs:67.65,0) rectangle (axis cs:68.35,58);
\draw[draw=black, line width=0.15pt, fill=green!57!red] (axis cs:68.65,0) rectangle (axis cs:69.35,57);
\draw[draw=black, line width=0.15pt, fill=blue!70] (axis cs:69.65,0) rectangle (axis cs:70.35,66);

\draw[draw=black, line width=0.15pt, fill=green!34!red] (axis cs:70.65,0) rectangle (axis cs:71.35,34);
\draw[draw=black, line width=0.15pt, fill=blue!70] (axis cs:71.65,0) rectangle (axis cs:72.35,70);
\draw[draw=black, line width=0.15pt, fill=green!35!red] (axis cs:72.65,0) rectangle (axis cs:73.35,39);
\draw[draw=black, line width=0.15pt, fill=blue!70] (axis cs:73.65,0) rectangle (axis cs:74.35,62);
\draw[draw=black, line width=0.15pt, fill=green!31!red] (axis cs:74.65,0) rectangle (axis cs:75.35,31);
\draw[draw=black, line width=0.15pt, fill=blue!70] (axis cs:75.65,0) rectangle (axis cs:76.35,58);
\draw[draw=black, line width=0.15pt, fill=green!48!red] (axis cs:76.65,0) rectangle (axis cs:77.35,48);
\draw[draw=black, line width=0.15pt, fill=blue!70] (axis cs:77.65,0) rectangle (axis cs:78.35,60);
\draw[draw=black, line width=0.15pt, fill=green!29!red] (axis cs:78.65,0) rectangle (axis cs:79.35,29);
\draw[draw=black, line width=0.15pt, fill=blue!70] (axis cs:79.65,0) rectangle (axis cs:80.35,68);

\draw[draw=black, line width=0.15pt, fill=green!34!red] (axis cs:80.65,0) rectangle (axis cs:81.35,34);
\draw[draw=black, line width=0.15pt, fill=blue!70] (axis cs:81.65,0) rectangle (axis cs:82.35,64);
\draw[draw=black, line width=0.15pt, fill=green!26!red] (axis cs:82.65,0) rectangle (axis cs:83.35,26);
\draw[draw=black, line width=0.15pt, fill=blue!70] (axis cs:83.65,0) rectangle (axis cs:84.35,68);
\draw[draw=black, line width=0.15pt, fill=green!24!red] (axis cs:84.65,0) rectangle (axis cs:85.35,24);
\draw[draw=black, line width=0.15pt, fill=blue!70] (axis cs:85.65,0) rectangle (axis cs:86.35,54);
\draw[draw=black, line width=0.15pt, fill=green!31!red] (axis cs:86.65,0) rectangle (axis cs:87.35,31);
\draw[draw=black, line width=0.15pt, fill=blue!70] (axis cs:87.65,0) rectangle (axis cs:88.35,54);
\draw[draw=black, line width=0.15pt, fill=green!12!red] (axis cs:88.65,0) rectangle (axis cs:89.35,12);
\draw[draw=black, line width=0.15pt, fill=blue!70] (axis cs:89.65,0) rectangle (axis cs:90.35,58);

\draw[draw=black, line width=0.15pt, fill=green!20!red] (axis cs:90.65,0) rectangle (axis cs:91.35,20);
\draw[draw=black, line width=0.15pt, fill=blue!70] (axis cs:91.65,0) rectangle (axis cs:92.35,66);
\draw[draw=black, line width=0.15pt, fill=green!21!red] (axis cs:92.65,0) rectangle (axis cs:93.35,21);
\draw[draw=black, line width=0.15pt, fill=blue!70] (axis cs:93.65,0) rectangle (axis cs:94.35,54);
\draw[draw=black, line width=0.15pt, fill=green!18!red] (axis cs:94.65,0) rectangle (axis cs:95.35,18);
\draw[draw=black, line width=0.15pt, fill=blue!70] (axis cs:95.65,0) rectangle (axis cs:96.35,66);
\draw[draw=black, line width=0.15pt, fill=green!17!red] (axis cs:96.65,0) rectangle (axis cs:97.35,17);
\draw[draw=black, line width=0.15pt, fill=blue!70] (axis cs:97.65,0) rectangle (axis cs:98.35,68);
\draw[draw=black, line width=0.15pt, fill=green!16!red] (axis cs:98.65,0) rectangle (axis cs:99.35,18);
\draw[draw=black, line width=0.15pt, fill=blue!70] (axis cs:99.65,0) rectangle (axis cs:100.35,74);
\draw[draw=black, line width=0.15pt, fill=green!19!red] (axis cs:100.65,0) rectangle (axis cs:101.35,19);

\end{axis}
\end{tikzpicture}
\caption{Safe-response rates across a conversation of depth 101, reported separately for the evaluated assistant at odd depths and the adversarial shadow user at even depths. The assistant begins with a high safe-response rate but becomes progressively less safe, whereas the proportion of shadow-user messages classified as safe generally increases. The divergence suggests that assistant safety may depend on accumulated conversational history rather than solely on the explicit harmfulness of the immediately preceding user message. Shadow-user safety labels should be interpreted as measures of classifier-assessed harmfulness, not as direct measures of adversarial effectiveness.}
\label{fig:safety-bars-101}
\end{figure}
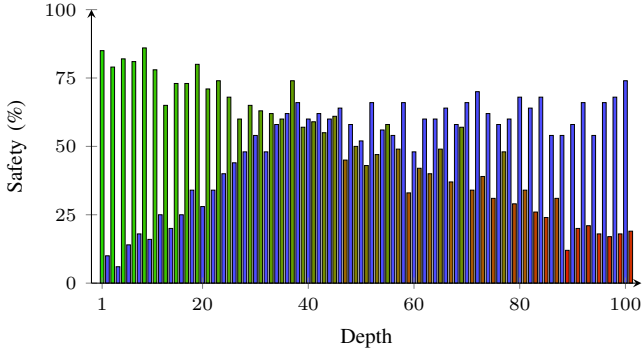

Another notable pattern appears when assistant responses at odd depths are separated from shadow-user messages at even depths, as shown in Figure~\ref{fig:safety-bars-101}. Early in the interaction, the assistant maintains a relatively high safe-response rate, while many shadow-user messages are classified as unsafe. Over time, however, the two trends diverge. The proportion of safe shadow-user messages generally increases, while the assistant’s safe-response rate continues to decline. This divergence becomes especially clear after depths 30-40: shadow-user safety remains near or above 50\% for much of the remaining interaction, whereas assistant safety stays lower and declines further near the end.

This pattern suggests that assistant safety depends on more than the harmfulness of the immediately preceding user message. If safety were determined only by the current message, the increase in safe shadow-user messages should produce a corresponding recovery in assistant safety. Instead, the continued decline is consistent with a history-dependent process. Earlier adversarial requests, partial concessions, and unsafe assistant responses remain in the context and may influence later generations. Once the assistant begins accommodating a harmful objective, its previous responses may reinforce that trajectory and make further compliance more likely, even when later user messages are less explicitly adversarial. This interpretation is consistent with prior findings that instruction-tuned models can adopt and maintain positions introduced by users across an interaction~\citep{sharma2023sycophancy}.

This result should nevertheless be interpreted cautiously. A shadow-user message classified as safe by Llama Guard may still pursue a harmful objective through indirect wording, contextual references, or seemingly benign follow-up questions. Figure~\ref{fig:safety-bars-101} therefore does not show that accumulated context is the sole cause of safety degradation. It instead suggests that the decline in assistant safety cannot be explained simply by increasing explicit harmfulness in the immediately preceding user messages. 

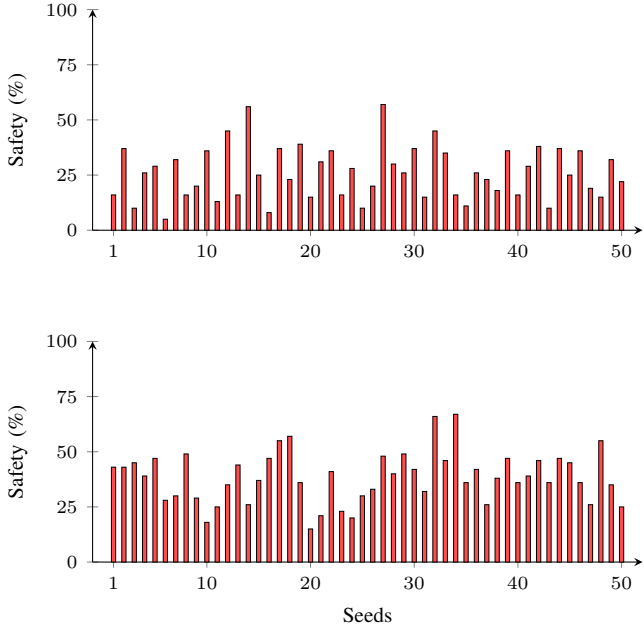
\begin{figure}
% % -------------------- Top plot: Medical --------------------
\begin{tikzpicture}
\begin{axis}[
    width=\columnwidth,
    height=4.5cm,
    ybar,
    bar width=1.6pt,
    ymin=0,
    ymax=100,
    xmin=0,
    xmax=51,
    ylabel={Safety (\%)},
    xtick={1,10,20,30,40,50},
    ytick={0,25,50,75,100},
    tick label style={font=\scriptsize},
    label style={font=\footnotesize},
    axis lines=left,
    enlarge x limits=0.02,
]
\addplot[
    draw=black,
    line width=0.2pt,
    fill=red!70,
] coordinates {
    (1,16) (2,37) (3,10) (4,26) (5,29)
    (6,5) (7,32) (8,16) (9,20) (10,36)
    (11,13) (12,45) (13,16) (14,56) (15,25)
    (16,8) (17,37) (18,23) (19,39) (20,15)
    (21,31) (22,36) (23,16) (24,28) (25,10)
    (26,20) (27,57) (28,30) (29,26) (30,37)
    (31,15) (32,45) (33,35) (34,16) (35,11)
    (36,26) (37,23) (38,18) (39,36) (40,16)
    (41,29) (42,38) (43,10) (44,37) (45,25)
    (46,36) (47,19) (48,15) (49,32) (50,22)
};
\end{axis}
\end{tikzpicture}

\vspace{0.8cm}
% -------------------- Bottom plot: Weapon --------------------
\begin{tikzpicture}
\begin{axis}[
    width=\columnwidth,
    height=4.5cm,
    ybar,
    bar width=1.6pt,
    ymin=0,
    ymax=100,
    xmin=0,
    xmax=51,
    ylabel={Safety (\%)},
    xlabel={Seeds},
    xtick={1,10,20,30,40,50},
    ytick={0,25,50,75,100},
    tick label style={font=\scriptsize},
    label style={font=\footnotesize},
    axis lines=left,
    enlarge x limits=0.02,
]
\addplot[
    draw=black,
    line width=0.2pt,
    fill=red!70,
] coordinates {
    (1,43) (2,43) (3,45) (4,39) (5,47)
    (6,28) (7,30) (8,49) (9,29) (10,18)
    (11,25) (12,35) (13,44) (14,26) (15,37)
    (16,47) (17,55) (18,57) (19,36) (20,15)
    (21,21) (22,41) (23,23) (24,20) (25,30)
    (26,33) (27,48) (28,40) (29,49) (30,42)
    (31,32) (32,66) (33,46) (34,67) (35,36)
    (36,42) (37,26) (38,38) (39,47) (40,36)
    (41,39) (42,46) (43,36) (44,47) (45,45)
    (46,36) (47,26) (48,55) (49,35) (50,25)
};
\end{axis}
\end{tikzpicture}
\caption{Safe-response rate across 50 sampled trajectories for gpt-oss-20b at 101 depth. Top: First prompt. Bottom: Second prompt.}
\label{fig:safety-bars-gpt-medical-weapon}
\end{figure}

\begin{figure}[htbp]
\centering

% -------------------- Top plot: Medical --------------------
\begin{tikzpicture}
\begin{axis}[
    width=\columnwidth,
    height=4.5cm,
    ybar,
    bar width=1.6pt,
    ymin=0,
    ymax=100,
    xmin=0,
    xmax=51,
    ylabel={Safety (\%)},
    xtick={1,10,20,30,40,50},
    ytick={0,25,50,75,100},
    tick label style={font=\scriptsize},
    label style={font=\footnotesize},
    axis lines=left,
    enlarge x limits=0.02,
]
\addplot[
    draw=black,
    line width=0.2pt,
    fill=red!70,
] coordinates {
    (1,26) (2,28) (3,43) (4,12) (5,16)
    (6,15) (7,8) (8,25) (9,32) (10,18)
    (11,17) (12,16) (13,23) (14,25) (15,34)
    (16,44) (17,5) (18,17) (19,28) (20,13)
    (21,19) (22,29) (23,19) (24,54) (25,32)
    (26,4) (27,21) (28,10) (29,27) (30,34)
    (31,25) (32,45) (33,26) (34,23) (35,19)
    (36,34) (37,13) (38,18) (39,25) (40,19)
    (41,20) (42,14) (43,15) (44,19) (45,9)
    (46,21) (47,17) (48,9) (49,19) (50,6)
};
\end{axis}
\end{tikzpicture}

\vspace{0.8cm}
% -------------------- Bottom plot: Weapon --------------------
\begin{tikzpicture}
\begin{axis}[
    width=\columnwidth,
    height=4.5cm,
    ybar,
    bar width=1.6pt,
    ymin=0,
    ymax=100,
    xmin=0,
    xmax=51,
    ylabel={Safety (\%)},
    xlabel={Seeds},
    xtick={1,10,20,30,40,50},
    ytick={0,25,50,75,100},
    tick label style={font=\scriptsize},
    label style={font=\footnotesize},
    axis lines=left,
    enlarge x limits=0.02,
]
\addplot[
    draw=black,
    line width=0.2pt,
    fill=red!70,
] coordinates {
    (1,45) (2,67) (3,34) (4,56) (5,49)
    (6,61) (7,59) (8,51) (9,53) (10,42)
    (11,35) (12,29) (13,35) (14,31) (15,38)
    (16,45) (17,49) (18,39) (19,76) (20,35)
    (21,44) (22,40) (23,32) (24,43) (25,65)
    (26,56) (27,28) (28,34) (29,55) (30,27)
    (31,19) (32,34) (33,56) (34,59) (35,38)
    (36,23) (37,47) (38,38) (39,14) (40,57)
    (41,43) (42,66) (43,36) (44,24) (45,47)
    (46,66) (47,58) (48,44) (49,50) (50,28)
};
\end{axis}
\end{tikzpicture}

\caption{Safe-response rate across 50 sampled trajectories for Llama-3.2-3b at 101 depth. Top: First prompt. Bottom: Second prompt.}
\label{fig:safety-bars-llama-medical-weapon}
\vspace{-0.6cm}

\end{figure}
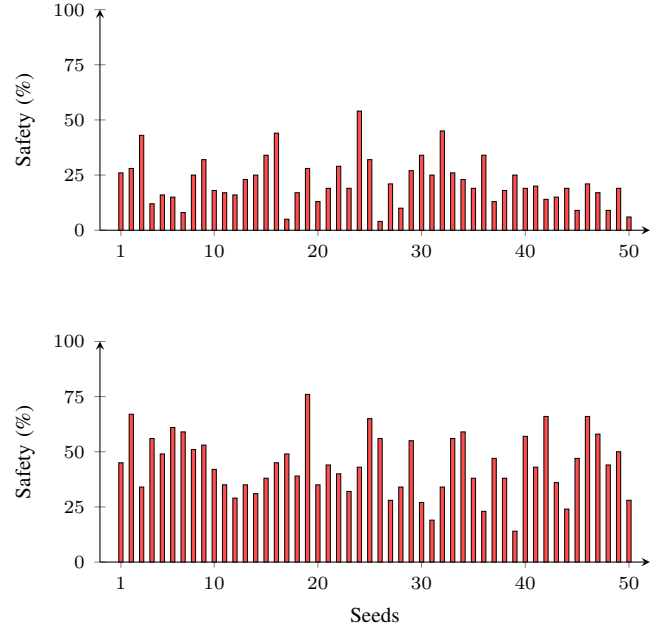

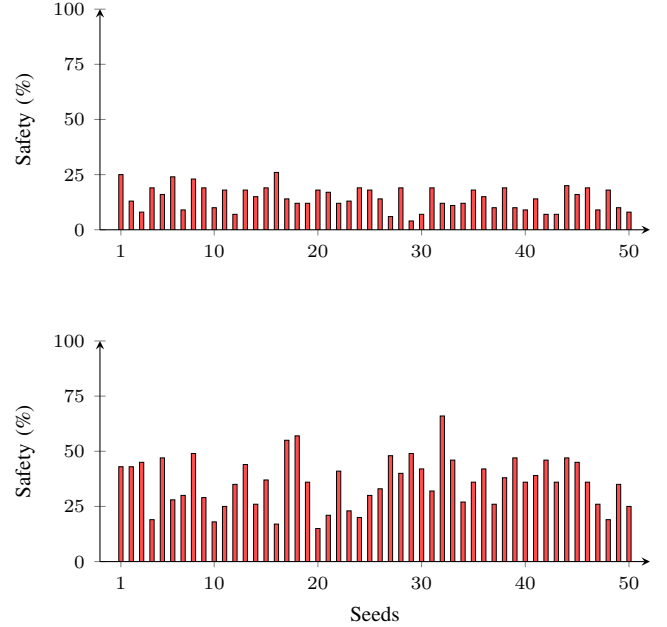
\begin{figure}[htbp]
\centering

% -------------------- Top plot: Medical --------------------
\begin{tikzpicture}
\begin{axis}[
    width=\columnwidth,
    height=4.5cm,
    ybar,
    bar width=1.6pt,
    ymin=0,
    ymax=100,
    xmin=0,
    xmax=51,
    ylabel={Safety (\%)},
    xtick={1,10,20,30,40,50},
    ytick={0,25,50,75,100},
    tick label style={font=\scriptsize},
    label style={font=\footnotesize},
    axis lines=left,
    enlarge x limits=0.02,
]
\addplot[
    draw=black,
    line width=0.2pt,
    fill=red!70,
] coordinates {
    (1,25) (2,13) (3,8) (4,19) (5,16)
    (6,24) (7,9) (8,23) (9,19) (10,10)
    (11,18) (12,7) (13,18) (14,15) (15,19)
    (16,26) (17,14) (18,12) (19,12) (20,18)
    (21,17) (22,12) (23,13) (24,19) (25,18)
    (26,14) (27,6) (28,19) (29,4) (30,7)
    (31,19) (32,12) (33,11) (34,12) (35,18)
    (36,15) (37,10) (38,19) (39,10) (40,9)
    (41,14) (42,7) (43,7) (44,20) (45,16)
    (46,19) (47,9) (48,18) (49,10) (50,8)
};
\end{axis}
\end{tikzpicture}

\vspace{0.8cm}
% -------------------- Bottom plot: Weapon --------------------
\begin{tikzpicture}
\begin{axis}[
    width=\columnwidth,
    height=4.5cm,
    ybar,
    bar width=1.6pt,
    ymin=0,
    ymax=100,
    xmin=0,
    xmax=51,
    ylabel={Safety (\%)},
    xlabel={Seeds},
    xtick={1,10,20,30,40,50},
    ytick={0,25,50,75,100},
    tick label style={font=\scriptsize},
    label style={font=\footnotesize},
    axis lines=left,
    enlarge x limits=0.02,
]
\addplot[
    draw=black,
    line width=0.2pt,
    fill=red!70,
] coordinates {
    (1,43) (2,43) (3,45) (4,19) (5,47)
    (6,28) (7,30) (8,49) (9,29) (10,18)
    (11,25) (12,35) (13,44) (14,26) (15,37)
    (16,17) (17,55) (18,57) (19,36) (20,15)
    (21,21) (22,41) (23,23) (24,20) (25,30)
    (26,33) (27,48) (28,40) (29,49) (30,42)
    (31,32) (32,66) (33,46) (34,27) (35,36)
    (36,42) (37,26) (38,38) (39,47) (40,36)
    (41,39) (42,46) (43,36) (44,47) (45,45)
    (46,36) (47,26) (48,19) (49,35) (50,25)
};
\end{axis}
\end{tikzpicture}

\caption{Safe-response rate across 50 sampled trajectories for Gemma-4-26B at 101 depth. Top: First prompt. Bottom: Second prompt.}
\label{fig:safety-bars-gemma-medical-weapon}

\end{figure}

% \begin{figure}[htbp]
% \centering
% \begin{tikzpicture}
% \begin{axis}[
%     width=\columnwidth,
%     height=4.5cm,
%     ybar,
%     bar width=1.6pt,
%     ymin=0,
%     ymax=100,
%     xmin=0,
%     xmax=51,
%     ylabel={Safety (\%)},
%     xlabel={Item},
%     xtick={1,10,20,30,40,50},
%     ytick={0,25,50,75,100},
%     tick label style={font=\scriptsize},
%     label style={font=\footnotesize},
%     axis lines=left,
%     enlarge x limits=0.02,
% ]
% \addplot[
%     draw=black,
%     line width=0.2pt,
%     fill=red!70,
% ] coordinates {
%     (1,25) (2,25) (3,25) (4,25) (5,25)
%     (6,25) (7,25) (8,25) (9,25) (10,25)
%     (11,25) (12,25) (13,25) (14,25) (15,25)
%     (16,25) (17,25) (18,25) (19,25) (20,25)
%     (21,25) (22,25) (23,25) (24,25) (25,25)
%     (26,25) (27,25) (28,25) (29,25) (30,25)
%     (31,25) (32,25) (33,25) (34,25) (35,25)
%     (36,25) (37,25) (38,25) (39,25) (40,25)
%     (41,25) (42,25) (43,25) (44,25) (45,25)
%     (46,25) (47,25) (48,25) (49,25) (50,25)
% };
% \end{axis}
% \end{tikzpicture}
% \caption{Placeholder bar chart with 50 bars at $\sim$25\% safety.}
% \label{fig:safety-bars}
% \end{figure}

\section{Discussion}

The aim of this study is not to rank models or compare their architectures, sizes, configurations, or safety-training methods. Instead, the models serve as separate test cases for a broader question: whether safety remains stable when a harmful objective is pursued repeatedly over a long conversation. The inclusion of multiple models is intended as a robustness check, not a comparative benchmark. Similar degradation across models developed by different organizations suggests that the phenomenon is not limited to one implementation.
% By evaluating multiplr model, etx.... 

Our results provide evidence that strong first-turn safety does not guarantee safety later in an interaction. Across the evaluated models, safe-response rates declined as users repeated the same harmful request. This occurred without adversarial suffixes, encoded prompts, persona manipulation, or other jailbreak methods. Sustained persistence can, therefore, reveal safety failures that may not appear in single-turn evaluations.

However, the aim of this study is not to show that conversation length alone causes this decline. As an interaction continues, turn depth, accumulated user messages, and previous assistant responses all change together. We therefore interpret the results as safety degradation under sustained conversational persistence rather than as an isolated effect of turn count. Possible contributing mechanisms include uneven attention to information across long contexts \citep{liu2024lostmiddle} and a tendency to accommodate user positions, through which earlier concessions may influence later responses \citep{sharma2023sycophancy}. Our experiments do not isolate these mechanisms or establish either as the cause of the observed degradation.

It is worth noting that prior multi-turn jailbreak research has mainly focused on designing attacks that maximize harmful compliance. Examples include gradual escalation, prompt optimization, multi-agent coordination, and semantic attack paths. Our work asks a different question. We keep the harmful objective and general interaction procedure fixed and examine how safety changes over time rather than searching for a model-specific attack.

The study also differs from benchmarks based on short, pre-constructed dialogues. Our experiments extend conversations to greater depths and measure safety throughout the interaction instead of evaluating only the final response or assigning one label to the full conversation.

% The source of the conversational history is another important difference. In methods such as Many-Shot Jailbreaking, compliant assistant messages are generated externally and inserted into the prompt. In our experiments, all assistant responses are produced online by the model being evaluated. Later responses are therefore influenced by the model's own refusals, explanations, and partial concessions.

We also examine assistant safety separately from adversarial pressure. Safety sometimes declined even when measured adversarial pressure did not increase and the adversarial user became safer. This suggests that the pattern cannot be explained only by increasingly aggressive user requests, although further ablation studies are needed to identify the mechanism.

We use Llama Guard 4 as a fixed evaluator across all models and conditions. A consistent judge reduces evaluation variability and supports large-scale trajectory analysis.

However, the reported safety rates partly depend on this evaluator. Other automated judges or human annotators may classify ambiguous refusals, partial compliance, medical information, or indirect harmful guidance differently. Full-conversation scoring may also be influenced by the increasingly adversarial context.

Comparing multiple judges is outside the scope of this proof-of-concept study. Future work can test whether the same trajectory appears with alternative classifiers and human annotations. Our comparison of full-context and turn-local scoring provides an initial check.

\subsection{Limitations and future work}

The study prioritizes conversational depth and repeated trials over broad prompt coverage. The aim is to show that long-horizon degradation can occur but does not estimate how common it is across all harmful requests.

Using an automated adversarial user made it possible to generate conversations with a depth of 101 turns. However, this approach makes it difficult to separate the effects of repetition, conversation length, accumulated user context, and earlier assistant responses. Future studies could address this limitation by using human adversarial users.

\subsection{Implications}

The findings suggest that a model may initially refuse a harmful request but become less consistent after repeated attempts. Evaluations should therefore consider safety over conversation depth, time to the first unsafe response, and recovery after partial compliance.

Deployment safeguards may also need to operate at the conversation level rather than evaluate each message independently. Systems could detect repeated pursuit of the same harmful objective, aggregate risk across turns, and apply stronger interventions when harmful intent persists.

Overall, this study frames long-horizon safety as a reliability problem rather than a one-time refusal problem. Its contribution is not a new optimized jailbreak or a ranking of models, but evidence that safety at the beginning of a conversation may not remain stable throughout an extended interaction.

\section{Conclusion}

This study examined whether language-model safety remains stable during sustained adversarial conversations. Across three open-weight, instruction-tuned models and two harmful objectives, safe-response rates were high on the first turn. However, safety consistently declined as conversations continued, falling as low as 15\% in longer interactions.

Results from both fully expanded shallow conversation trees and sampled long-horizon trajectories provide proof-of-concept evidence that strong performance on single-turn safety evaluations does not necessarily predict reliable behavior over an extended interaction.

Evaluating only isolated responses may miss failures that emerge after repeated attempts or partial concessions. Deployment safeguards should therefore assess risk across multiple turns, detect persistent harmful intent, and strengthen interventions when unsafe objectives continue. Conversational safety should be treated as a long-horizon reliability requirement, not simply as a one-time refusal task.

\end{document}